\documentclass[10pt,conference]{IEEEtran}

\usepackage{cite}

\ifCLASSINFOpdf
   \usepackage[pdftex]{graphicx}
   \graphicspath{{figs/}}
   \DeclareGraphicsExtensions{.pdf,.jpeg,.png}
\else
   \usepackage[dvips]{graphicx}
   \graphicspath{{../figs/}}
   \DeclareGraphicsExtensions{.eps}
\fi
\usepackage[cmex10]{amsmath}
\usepackage{amsthm}

\usepackage{algorithmic}

\usepackage{array}

\ifCLASSOPTIONcompsoc
  \usepackage[caption=false,font=normalsize,labelfont=sf,textfont=sf]{subfig}
\else
  \usepackage[caption=false,font=footnotesize]{subfig}
\fi
\usepackage{url}

\usepackage{multirow}
\usepackage{adjustbox}
\usepackage{xcolor}
\usepackage{pgfplots}
\pgfplotsset{compat=1.17}

\newif\iffinal
\finaltrue
\newcommand{\cmtid}{79}

\iffinal
\else
\usepackage[switch]{lineno}
\fi

\begin{document}
%
\title{Iterative Pseudo-Labeling with Deep Feature Annotation and Confidence-Based Sampling}


\iffinal

\author{
    \IEEEauthorblockN{Bárbara C. Benato\IEEEauthorrefmark{1}, Alexandru C. Telea\IEEEauthorrefmark{2}, Alexandre X. Falcão\IEEEauthorrefmark{1}}
    \IEEEauthorblockA{\IEEEauthorrefmark{1}Laboratory of Image Data Science, \\Institute of Computing, University of Campinas, \\Campinas, Brazil
    \\\{barbara.benato, afalcao\}@ic.unicamp.br}
    \IEEEauthorblockA{\IEEEauthorrefmark{2}
    Department of Information and Computing Sciences, \\Faculty of Science, Utrecht University, \\Utrecht, The Netherlands
    \\a.c.telea@uu.nl}
}

\else
  \author{Sibgrapi paper ID: \cmtid \\ }
  \linenumbers
\fi

\maketitle

\begin{abstract}
Training deep neural networks is challenging when large and annotated datasets are unavailable. Extensive manual annotation of data samples is time-consuming, expensive, and error-prone, notably  when it needs to be done by experts. To address this issue, increased attention has been devoted to techniques that propagate uncertain labels (also called pseudo labels) to large amounts of unsupervised samples and use them for training the model. However, these techniques still need hundreds of supervised samples per class in the training set and a validation set with extra supervised samples to tune the model. We improve a recent iterative pseudo-labeling technique, \emph{Deep Feature Annotation} (DeepFA), by selecting the most confident unsupervised samples to iteratively train a deep neural network. Our confidence-based sampling strategy relies on only \emph{dozens} of annotated training samples per class with no validation set, considerably reducing user effort in data annotation. We first ascertain the best configuration for the baseline -- a self-trained deep neural network -- and then evaluate our \textit{confidence} DeepFA for different confidence thresholds. Experiments on six datasets show that DeepFA already outperforms the self-trained baseline, but confidence DeepFA can considerably outperform the original DeepFA and the baseline. 
\end{abstract}

\IEEEpeerreviewmaketitle

\section{Introduction}
The success of supervised deep neural networks is evident in many applications.
However, the need for large annotated training sets is a well-known problem\,\cite{Lin:2014,Sun:2017}. 
Data augmentation and transfer learning aim to address this problem, with semi-supervised learning\,\cite{Zhun:2018, Wu:2018,Iscen:2019}, its variants pseudo-labeling\,\cite{Lee:2013, Jing:2020} and meta pseudo-labeling\,\cite{Pham:2021:CVPR}, and few-shot learning\,\cite{Sung:2018,Sun:2019}, all receiving increased attention. 

Semi-supervised learning methods\,\cite{Zhun:2018, Wu:2018,Iscen:2019} propagate labels from a small set of supervised samples to a large set of unsupervised ones by exploiting their distribution in a given latent feature space. Pseudo-labeling approaches\,\cite{Lee:2013, Jing:2020} (a particular case of self-training) essentially adopt the semi-supervised strategy with the apprentice model assigning uncertain (pseudo) labels to unsupervised samples. Meta pseudo-labeling\,\cite{Pham:2021:CVPR} uses an auxiliary model (teacher) to generate pseudo labels to train the primary model (student). In few-shot learning\,\cite{Sung:2018,Sun:2019}, the model is designed from a handful of supervised samples with or without unlabeled data. Whenever many unsupervised samples are available, semi-supervised learning techniques should be preferred to increase the number of labeled training samples and, consequently, improve feature learning and classification performances.  

In semi-supervised learning, pseudo-labeling was first proposed for more effectively fine-tuning a pretrained model\,\cite{Lee:2013}. The model can be retrained with a large annotated dataset by assuming that pseudo labels are actual labels. Yet, label propagation errors can negatively affect the performance of classifiers trained from them\,\cite{BenatoSibgrapi:2018, Arazo:2020}. To mitigate the problem, the confidence of the apprentice model has been included in the loss function\,\cite{Iscen:2019, Shi:2018}. Yet, pseudo-labeling methods still require a training set with hundreds of supervised samples per class and a validation set with extra supervised samples (at least other 1000 labeled samples) to guarantee reasonable label-propagation accuracy~\cite{Lee:2013,Miyato:2018,Jing:2020, Pham:2021:CVPR}.

In this work, we improve an iterative meta pseudo-labeling strategy, named \emph{Deep Feature Annotation} (\emph{DeepFA})\,\cite{Benato:2021}, using the confidence of the auxiliary model used for label propagation to select the unsupervised samples for training the primary model. In \emph{DeepFA}, the auxiliary model is a combination of the t-SNE projection technique\,\cite{MaatenJMLR:14} applied to a latent feature space (the last convolutional layer) of the primary model, and a semi-supervised optimum-path forest (OPFSemi) classifier\,\cite{Amorim:2016}. The primary model is trained with a small supervised set, generating the first latent feature set for label propagation by the auxiliary model. In DeepFA, the auxiliary model propagates labels to all unsupervised samples, similar to\,\cite{Amorim:2019}. This may increase label propagation errors since unsupervised samples, which are far away from supervised ones, may receive incorrect labels. As OPFSemi has no parameters to optimize, DeepFA does not need an extra validation set with supervised samples.

In contrast, we propose to retrain the primary model by using only the most confident unsupervised samples, with confidence given by the label propagation method. We call this variant \textit{confidence DeepFA} (\emph{conf-DeepFA}). Retraining the primary model, latent feature projection, and label propagation repeat for a few iterations to improve the primary model. At each iteration, the primary model is expected to improve its latent feature space reducing the label propagation errors of the auxiliary model. Given that OPFSemi is sensitive to the curse of high dimensionality, we use t-SNE to reduce the feature space to two dimensions. This combination produces a labeling function statistically independent to the labeling function of the primary model, satisfying the main requirement for meta-learning. 

We first assess the best training scheme of the primary model and use it as a baseline. We selected VGG-16\,\cite{VGG16} pretrained with ImageNet and adapted it for six different datasets with and without iterative pseudo-labeling. We then compared the best scheme against the original \emph{DeepFA}\,\cite{Benato:2021} and our \emph{conf-DeepFA} using both fixed and adaptive thresholds to select samples with the most confident labels. To emphasize the potential of  \emph{conf-DeepFA}, we used only $1\%$ of supervised samples (i.e., just dozens of samples per class) and no validation set. Our results indicate that our method can significantly outperform the original DeepFA and the baseline methods.

\section{Deep Feature Annotation}
\begin{figure*}[hbt] 
    \centering 
        \includegraphics[width=5in]{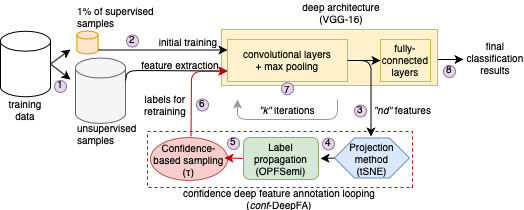} 
    \caption{Proposed \emph{conf-DeepFA} method. 
    A training dataset is split into (1)  supervised and unsupervised sets. We consider only $1\%$ of supervised samples (images) to train a deep neural network (2). Using the network, we extract features from both supervised and unsupervised data (3) and project them in a 2D embedded space (4). In this space, we propagate labels from the supervised samples (5). However, we select only unsupervised samples with the most confident labels, as assigned by the label propagation method (6), to retrain the deep neural network (7). The loop 3-4-5-6-7 repeats for a few iterations. Finally, the  classification results (8) are obtained from the fully-connected layers of the trained model. Our contribution extends \emph{DeepFA}\,\cite{Benato:2020:arxiv} by adding the confidence-based sampling (red round shape in the red-dashed box).}
    \label{f.pipeline}
\end{figure*}

As a semi-supervised method, \emph{DeepFA}~\cite{Benato:2020:arxiv} iteratively repeats three steps -- deep feature learning, feature space projection, and pseudo-labeling -- as described next (see also Fig.~\ref{f.pipeline}).

\subsection{Deep Feature Learning}
One may conceptually divide a classification deep neural network into (a) layers for feature extraction, (b) fully connected layers for feature space reduction and (c) a decision layer for prediction, being (b-c) an MLP classifier. We are interested in the features of the last convolutional layer of VGG-16 (after max-pooling) that result from (a), where the feature space is still high and sparse.

To minimize user effort for annotation, \emph{DeepFA} uses the ability of pre-trained CNNs to transfer knowledge between scenarios – e.g., from natural to medical images – using few supervised samples and few training epochs. To do this, in our work, we fine-tune VGG-16 with ImageNet's\,\cite{imagenet:2015} pre-trained weights using the few available supervised images. 
Finally, the true-labeled images and pseudo-labeled ones by \emph{DeepFA} are used to retrain VGG-16 in the next iterations of our loop.

\subsection{Feature Space Projection}
The features of VGG-16's last convolutional layer are projected by t-SNE\,\cite{MaatenJMLR:14} on a 2D embedded space. Rauber et al.\,\cite{RauberInfVis2017} showed that high classification accuracy relates to a good separation of classes in a 2D projection of the samples' feature vectors. More exactly, they showed that, if a 2D projection (in particular, t-SNE) presents good class separation, then a good class separation can also be found in the data space. 
Benato et al.\,\cite{BenatoSibgrapi:2018, Benato:2021} showed that label propagation (using two semi-supervised classifiers) in a 2D t-SNE projection space leads to better classification results than label propagation in the higher-dimensional latent feature space of an autoencoder. Extending the above, Benato et al.\,\cite{Benato:2020:arxiv} used label propagation in a 2D projected space to create large training sets for deep learning. 

\subsection{Pseudo-labeling}
OPFSemi was used for pseudo-labeling on a 2D t-SNE projection\,\cite{BenatoSibgrapi:2018,Benato:2021,Benato:2020:arxiv} and in the original high-dimensional feature space\,\cite{Amorim:2019,Benato:2021}. Although OPFSemi's confidence values have been used to improve pseudo-labeling for data annotation, this was not analyzed within the iterative pseudo-labeling loop in\,\cite{Benato:2020:arxiv} to create large training sets for deep learning.

OPFSemi maps supervised and unsupervised samples to nodes of a graph and computes an optimum-path forest rooted at supervised samples. Two types of cost values are calculated to each unsupervised sample. Let $L(u) \in {1,2,...,c}$ be the (pseudo) label assigned by OPFSemi to an unsupervised sample $u$. The label $L(u)$ is equal to the class $\lambda(s) \in {1,2,...,c}$ of the supervised sample $s$, which has offered the optimum path to $u$ among paths offered from all supervised samples. Let $C(u)$ be the cost of that optimum path from $s$ and $C'>C(u)$ be the second least path cost offered to $u$ by a supervised sample $p$, whose class $\lambda(p)$ is different to the class $L(u)=\lambda(s)$. Then the confidence value $V(u) = C' / (C(u) + C') \in [0,1]$ is assigned to the unsupervised sample $u$. Higher is $C'$ more confident OPFSemi is that $L(u)$ is $\lambda(s)$. 

All labels assigned by OPFSemi with a confidence $V$ above a threshold $\tau$ are used for VGG-16's training. In this work, we explore different $\tau$ values and propose an adaptive approach in which $\tau$ is increased along the \emph{conf-DeepFA} iterations. In contrast to~\cite{Benato:2021}, we do not ask the user to choose the $\tau$ value, since we wish to validate a confidence-based sampling based on OPFSemi during the \emph{DeepFA}'s looping. While this simplifies our approach by releasing the user from the effort of choosing a (good) $\tau$ threshold, this also potentially removes insights that users could use to select better $\tau$ values. Exploring how our automatic method compares to the user-in-the-loop approach is subject for future work.

\section{Experiments}

\subsection{Datasets}
\label{sec:datasets}
We choose six datasets to validate our proposed method. The first one is the public  MNIST\,\cite{Lecun:2010:mnist} dataset. MNIST has $0$ to $9$ handwritten digits grayscale images of $28 \times 28$ pixels. We use $5000$ random samples from the original training dataset. 

\begin{figure}[htb]
\centering
  \includegraphics[width=1.0\linewidth]{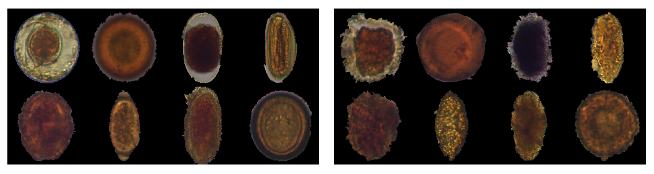}\\
  \caption{Examples of species of H.Eggs (left) and similar images of impurities (right).}
  \label{f.parasites}
\end{figure}

The subsequent five datasets come from a Parasite image collection\,\cite{Suzuki:2013}. This collection has three main dataset types: (i) \emph{Helminth larvae}, (ii) \textit{Helminth eggs}, and (iii) \emph{Protozoan cysts}. These datasets contain color microscopy images of $200 \times 200$ pixels of the most common species of human intestinal parasites in Brazil, responsible for public health problems in most tropical countries\,\cite{Suzuki:2013}. The datasets are challenging since they are unbalanced and contain an impurity class as the majority class, having samples very similar to parasites, making classification hard (see Fig.~\ref{f.parasites}).
Table~\ref{t.datasets} shows the number, type, and amount of samples per class for each of the three datasets (i-iii) listed above. To these three datasets, we also add the \emph{Helminth eggs} and  \emph{Protozoan cysts} datasets without the impurity class, leading to a total of 5 datasets. 

\begin{table}[th]
\caption{Three Parasites datasets: number of classes, class names, and number of samples per class.}
\label{t.datasets}
\centering
\renewcommand{\arraystretch}{0.9}
\begin{tabular}{l|l|r}
\multicolumn{1}{c|}{dataset}                                                           & \multicolumn{1}{c|}{class} & \multicolumn{1}{c}{$\#$ samples} \\ \hline
\multirow{3}{*}{\begin{tabular}[c]{@{}l@{}}(i) \textit{Helminth larvae}\\ (2 classes)\end{tabular}} & \textit{S.stercoralis}                     & 446                              \\
                                                                                       & impurities                 & 3068                              \\
                                                                                       &                           total & 3,514                   \\ \hline
\multirow{10}{*}{\begin{tabular}[c]{@{}l@{}}(ii) \textit{Helminth eggs}\\ (9 classes)\end{tabular}}  & \textit{H.nana}                     & 348                              \\
                                                                                       & \textit{H.diminuta}                 & 80                              \\
                                                                                       & \textit{Ancilostomideo}             & 148                              \\
                                                                                       & \textit{E.vermicularis}             & 122                              \\
                                                                                       & \textit{A.lumbricoides}             & 337                              \\
                                                                                       & \textit{T.trichiura}                & 375                              \\
                                                                                       & \textit{S.mansoni}                  & 122                              \\
                                                                                       & \textit{Taenia}                     & 236                              \\
                                                                                       & impurities                 & 3,444                              \\
                                                                                       &                          total  & 5,112                   \\ \hline
\multirow{8}{*}{\begin{tabular}[c]{@{}l@{}}(iii) \textit{Protozoan cysts}\\ (7 classes)\end{tabular}} & \textit{E.coli}                     & 719                              \\
                                                                                       & \textit{E.histolytica}              & 78                              \\
                                                                                       & \textit{E.nana}                     & 724                              \\
                                                                                       & \textit{Giardia}                    & 641                              \\
                                                                                       & \textit{I.butschlii}                & 1,501                              \\
                                                                                       & \textit{B.hominis}                  & 189                              \\
                                                                                       & impurities                 & 5,716                              \\
                                                                                       &                           total & 9,568                         
\end{tabular}
\end{table}

\subsection{Experimental Setup}
To reproduce the scenario of few supervised samples, we define a supervised training set $S$ with only $1\%$ of supervised samples of an entire dataset $D$, assuming the unsupervised $U$ and test $T$ sets with $69\%$ and $30\%$ of samples, respectively ($D = S \cup U \cup T$).
A very small $S$ simulates the real-world scenario when one has a large $D$ but manual effort is needed to label samples to create $S$. We randomly divide each dataset $D$ into $S$, $U$, and $T$ in a stratified manner and also generate \emph{three} distinct splits for each experiment for further statistical analysis of our classification results. Table~\ref{t.samples} shows the number of supervised samples in $S$ for each of the six datasets introduced in Sec.~\ref{sec:datasets}.
\begin{table}[th]
\caption{Number of supervised samples in $S$ for each chosen dataset.}
\label{t.samples}
\centering
\begin{adjustbox}{width=0.9\linewidth}
\begin{tabular}{c|c|c|c|c|c|c}
\multicolumn{1}{l|}{} & \begin{tabular}[c]{@{}c@{}}MNIST\end{tabular} & \begin{tabular}[c]{@{}c@{}}H.eggs\\ (w/o imp)\end{tabular} & \begin{tabular}[c]{@{}c@{}}P. cysts\\ (w/o imp)\end{tabular} & H. larvae & H. eggs & P. cysts \\ \hline
S                     & 50                                                       & 17                                                         & 38                                                           & 35        & 51      & 95       \\ \hline
U                     & 3450                                                     & 1220                                                       & 2658                                                         & 2424      & 3527    & 6602    
\end{tabular}
\end{adjustbox}
\end{table}

To evaluate our method, we get the probability of VGG-16's last fully-connected layer and compute accuracy and Cohen’s $\kappa$, since we have unbalanced datasets. $\kappa \in [-1, 1]$ gives the agreement level between two distinct predictions, where $\kappa \leq 0$ means no possibility and $\kappa = 1$ means the full possibility of agreement occurring by chance, respectively. We also compute the number of correct labels assigned in $U$ for each proposed experiment to evaluate the label propagation accuracy.

\subsection{Implementation details}
\label{sec:implem}
As stated before, our aim is using \emph{conf-DeepFA} without an additional validation set, whose creation, as explained, would ask for more user supervision (i.e., effort in data annotation). For this, we fix all pipeline's parameters without any parameter optimization step. Specifically: OPFSemi has no parameters; for t-SNE, we used the default parameters in\,\emph{scikit-learn}. 

The VGG-16 architecture was implemented in Python using Keras\,\cite{Chollet:2015:keras}. The original fully-connected layers were replaced by two fully connected layers with $4096$ neurons and rectified linear activation, followed by a decision layer with $c$ neurons, where $c$ equals the number of classes for each dataset (see Tab.~\ref{t.datasets}),
and softmax activation. The model is trained by error backpropagation for a categorical cross-entropy function and using stochastic gradient descent with a linearly decaying learning rate initialized at $0.1$ and momentum of $0.9$, respectively. We loaded ImageNet pre-trained weights and used a linear decay of $1\times10^{-6}$ over $15$ epochs. The pre-trained weights for convolutional layers were fixed for the feature extraction experiments and unfrozen for fine-tuning, respectively.

\subsection{Proposed experiments}
\label{ss.proposed_exp}
First, we evaluate the impact of VGG-16's training with and without fine-tuning the convolutional layers when loading ImageNet pre-trained weights (Sec.~\ref{sec:implem}). We also  evaluate VGG-16 for label propagation, using pseudolabels produced by VGG-16 to feed its training in the next iteration of the data annotation looping (see Fig.~\ref{f.pipeline}). We executed four experiments, as follows; in the next items, \emph{ft} stands for fine-tuning and \emph{fe} stands for feature extraction, respectively:

\begin{itemize}
    \item \emph{VGG-16$_{ft}$}: VGG-16 with pre-trained weights and convolutional layers \emph{unfrozen} trained on $S$ and tested on $T$; 
    \item \emph{self-VGG-16$_{ft}$}: VGG-16 with pre-trained weights and convolutional layers \emph{unfrozen} trained on $S$. Pseudolabels are obtained for \emph{all} samples in $U$. $S$ and the  pseudolabeled $U$ are used to train VGG-16, and the network is tested on $T$. Each one of the $5$ iterations repeats this process;
    \item \emph{VGG-16$_{fe}$}: VGG-16 with pre-trained weights and convolutional layers \emph{frozen} trained on $S$ and tested on $T$; 
    \item \emph{self-VGG-16$_{fe}$}: VGG-16 with pre-trained weights and convolutional layers \emph{frozen} trained on $S$. Pseudolabels are obtained for \emph{all} samples in  $U$. $S$ and $U$ are used to train VGG-16, and the network is tested on $T$ (each one of the $5$ iterations repeats this process).
\end{itemize}

We found out that self-VGG-16$_{fe}$ achieves better results (Sec.~\ref{ss.exps}), so we defined this training procedure for the subsequent experiments described below. We evaluate the impact of  OPFSemi's confidence sample in the \emph{DeepFA} looping by the following experiments:

\begin{itemize}
    \item \emph{DeepFA}: VGG-16 is trained on $S$. Deep features for $S\cup U$ from the last convolutional layer are projected in 2D with t-SNE, and used next for OPFSemi pseudo labeling from $S$ to \emph{all} samples in $U$. OPFSemi's pseudolabels are used to retrain VGG-16, and the network is tested on $T$ (one iteration of \emph{DeepFA} looping out of five); 
    \item \emph{conf-DeepFA$_{\tau = x}$}: VGG-16 is trained on $S$. Deep features for $S\cup U$ from the last convolutional layer are projected in 2D with t-SNE, and used for OPFSemi pseudo labeling from $S$ to $U_\tau$, for samples with confidence above $\tau=x$. We choose $x=\{0.7, 0.8, 0.9\}$. OPFSemi's pseudolabels are used to retrain VGG-16, and the network is tested on $T$ (one iteration of \emph{conf-DeepFA} looping out of five);
    \item \emph{conf-DeepFA$_{\tau = \alpha}$}: VGG-16 is trained on $S$. Deep features for $S\cup U$ from the last convolutional layer are projected in 2D with t-SNE, and used for OPFSemi pseudo labeling from $S$ to $U_\tau$ for samples with confidence above $\tau$. $\tau$ is increased from $0.8$ to $0.96$ by $0.4$ in each \emph{conf-DeepFA} looping iteration. OPFSemi's pseudolabels are used to retrain VGG-16, and the network is tested on $T$. The looping has five iterations.
\end{itemize}

\subsection{Experimental results}
\label{ss.exps}
Table~\ref{t.exp_vgg} shows the results of the experiments in Sec.~\ref{ss.proposed_exp} that investigate the impact of the parameters' fine-tuning in a pre-trained VGG-16 deep architecture when using a small amount of supervised samples to generate pseudo labels. We show mean propagation accuracy, classification accuracy, $\kappa$, and standard deviation of \emph{three} different splits for VGG-16 trained only with $S$ and tested on $T$ with feature extraction (\emph{VGG-16$_{fe}$}) and fine-tuning (\emph{VGG-16$_{ft}$}). Also, we present the results for VGG-16 trained with its labeled samples on $U$ over five iterations (\emph{self-VGG-16$_{fe}$} and \emph{self-VGG-16$_{ft}$}). We see first that both experiments -- feature extraction and fine-tuning -- do not show relevant gain in propagation accuracy or $\kappa$ along with the iterations. The results are even worse from \emph{VGG-16$_{ft}$} to \emph{self-VGG-16$_{ft}$}. In general, VGG's feature extraction results show an increase of almost $20\%$ in accuracy and $\kappa$ for most datasets. This shows that the results in~\cite{Benato:2020:arxiv} can get better when using feature extraction instead of fine-tuning, even though we use fewer training epochs in our work.
\begin{table*}[!h]
\renewcommand{\arraystretch}{0.9}
\caption{Results for VGG-16 considering feature extraction and fine-tuning. Best values per metric and dataset in bold.}
\label{t.exp_vgg}
\scriptsize
\centering
\begin{tabular}{l|l|c|c|c|c}
dataset                             & metric    & \emph{VGG-16$_{ft}$}                & \emph{self-VGG-16$_{ft}$}           & \emph{VGG-16$_{fe}$}                & \emph{self-VGG-16$_{fe}$}          \\ \hline
\multirow{3}{*}{MNIST}              & prop. acc &            -         &  0.447238 $\pm$ 0.146 & -                                & \textbf{0.586000                 $\pm$ 0.007}   \\  
                                    & acc       & \textbf{0.629555 $\pm$ 0.037} &  0.441334 $\pm$ 0.149 & 0.614444       $\pm$ 0.015       & 0.592222                 $\pm$ 0.020   \\  
                                    & kappa     & \textbf{0.588195 $\pm$ 0.041} &  0.378648 $\pm$ 0.166 & 0.571176       $\pm$ 0.017       & 0.546162                 $\pm$ 0.023   \\ \hline
\multirow{3}{*}{\shortstack[l]{H.eggs\\ (w/o imp)}}   & prop. acc &            -         &  \textbf{0.758825 $\pm$ 0.088} &  -                               & 0.744004                 $\pm$ 0.114  \\  
                                    & acc       & \textbf{0.790961 $\pm$ 0.050} &  0.779033 $\pm$ 0.095 & 0.738858       $\pm$ 0.054       & 0.774011                 $\pm$ 0.131   \\  
                                    & kappa     & \textbf{0.752807 $\pm$ 0.060} &  0.735591 $\pm$ 0.113 & 0.693278       $\pm$ 0.060       & 0.734030                 $\pm$ 0.153   \\ \hline
\multirow{3}{*}{\shortstack[l]{P.cysts \\ (w/o imp)}}  & prop. acc &            -         &  0.399481 $\pm$ 0.010 & -                                & \textbf{0.648739                 $\pm$ 0.111}   \\  
                                    & acc       & 0.561130 $\pm$ 0.093 &  0.400519 $\pm$ 0.011 & \textbf{0.736159       $\pm$ 0.027}       & 0.650230                 $\pm$ 0.101    \\  
                                    & kappa     & 0.324051 $\pm$ 0.175 &  0.020734 $\pm$ 0.021 & \textbf{0.626632       $\pm$ 0.039}       & 0.483706                 $\pm$ 0.170    \\ \hline
\multirow{3}{*}{H.larvae}           & prop. acc &            -         &  0.897384 $\pm$ 0.031& -                                & \textbf{0.912837                 $\pm$ 0.038}     \\  
                                    & acc       & 0.874566 $\pm$ 0.001 &  0.886572 $\pm$ 0.017& 0.893523       $\pm$ 0.017       & \textbf{0.908689                 $\pm$ 0.040}     \\  
                                    & kappa     & 0.021406 $\pm$ 0.019 &  0.174158 $\pm$ 0.208 & 0.256836       $\pm$ 0.203       & \textbf{0.385892                 $\pm$ 0.402}    \\ \hline
\multirow{3}{*}{H.eggs}             & prop. acc &            -         &  0.773803 $\pm$ 0.034 & -                                & \textbf{0.847308                 $\pm$ 0.018}    \\  
                                    & acc       & \textbf{0.858323 $\pm$ 0.013} &  0.775750 $\pm$ 0.034 & 0.848327       $\pm$ 0.017       & \textbf{0.850934                 $\pm$ 0.014}    \\  
                                    & kappa     & \textbf{0.734333 $\pm$ 0.019} &  0.519971 $\pm$ 0.114 & 0.713649       $\pm$ 0.030       & 0.714227                 $\pm$ 0.038    \\ \hline
\multirow{3}{*}{P.cysts}            & prop. acc &            -         &  0.730327 $\pm$ 0.022 & -                                & \textbf{0.817978                 $\pm$ 0.004}   \\  
                                    & acc       & 0.758853 $\pm$ 0.077 &  0.734239 $\pm$ 0.028 & 0.818182       $\pm$ 0.004       & \textbf{0.824800                 $\pm$ 0.011}   \\  
                                    & kappa     & 0.542967 $\pm$ 0.218 &  0.492070 $\pm$ 0.107 & 0.697633       $\pm$ 0.009       & \textbf{0.705397                 $\pm$ 0.022}   \\ \hline
\end{tabular}
\end{table*}
 
A separate interesting question raised in~\cite{Benato:2020:arxiv} is: How to improve the OPFSemi's pseudo labeling over the iterations? To answer this, we propose to use the confidence-based sampling as stated before. Table~\ref{t.exp_deepfa} shows the mean propagation accuracy, classification accuracy, $\kappa$, and standard deviation of the three splits for the proposed experiments, for the last of the five executed iterations. For all datasets (except \emph{P.cysts}), we see that selecting the most confident samples by OPFSemi during the \emph{DeepFA} looping \emph{improves} the pseudo-labeling results over the iterations. When using $\tau = 0.8$, MNIST, \emph{H.larvae}, and \emph{P.cysts} obtained the best results. For \emph{H.eggs} without impurities, $\tau = 0.7$ shows the best results. For \emph{P.cysts} without impurities, the results of $\tau = 0.9$ and $\tau = \alpha$ (adaptive) show the best (and similar) results. The proposed confidence-based looping annotation did not improve \emph{P.cysts} with impurities. However, this dataset is also the most challenging one: seven classes, more samples, and almost $60\%$ total samples are impurities. We conclude that confidence-based sampling shows clear added value in nearly all situations. However, we also note that selecting $\tau$ may depend on the dataset and its difficulty.
\begin{table*}[!h]
\renewcommand{\arraystretch}{0.9}
\caption{Results from the last iteration for proposed experiments with fully label propagation (\emph{DeepFA}), and confidence-based label propagation (\emph{conf-DeepFA}) with confidence higher than $\tau=0.7$, confidence higher than $\tau=0.8$, confidence higher than $\tau=0.9$, and adaptative confidence (from $0.80$ to $0.96$ over 5 iterations). Best values per dataset in bold.}
\label{t.exp_deepfa}
\scriptsize
\centering
\begin{tabular}{l|l|c|c|c|c|c}
dataset                             & metric    & \emph{DeepFA}                   & \emph{conf-DeepFA$_{\tau=0.7}$}         & \emph{conf-DeepFA $_{\tau=0.8}$}    & \emph{conf-DeepFA $_{\tau=0.9}$}         & \emph{conf-DeepFA $_{\tau=\alpha}$}         \\ \hline
\multirow{3}{*}{MNIST}              & prop. acc & 0.790000           $\pm$ 0.047 & 0.782286  $\pm$ 0.029 &                 \textbf{0.821714 $\pm$ 0.018} & 0.750000          $\pm$ 0.028 & 0.795429            $\pm$ 0.007 \\  
                                    & acc       & 0.797778           $\pm$ 0.049 & 0.788000  $\pm$ 0.030 &                 \textbf{0.822666 $\pm$ 0.022} & 0.740222          $\pm$ 0.032 & 0.651778            $\pm$ 0.062 \\  
                                    & kappa     & 0.775103           $\pm$ 0.054 & 0.764348  $\pm$ 0.034 &                 \textbf{0.802863 $\pm$ 0.024} & 0.710961          $\pm$ 0.036 & 0.612766            $\pm$ 0.069 \\ \hline
\multirow{3}{*}{\shortstack[l]{H.eggs\\ (w/o imp)}}   & prop. acc & \textbf{0.983293           $\pm$ 0.004} & 0.983832  $\pm$ 0.002 &                 0.974401 $\pm$ 0.020 & 0.981945          $\pm$ 0.003 & 0.983832            $\pm$ 0.004 \\  
                                    & acc       & 0.790961 $\pm$ 0.050 & \textbf{0.973007  $\pm$ 0.006} &                 0.971123 $\pm$ 0.013 & 0.938481          $\pm$ 0.056 & 0.806654            $\pm$ 0.126 \\  
                                    & kappa     & 0.752807 $\pm$ 0.060 & \textbf{0.968042  $\pm$ 0.007} &                 0.965848 $\pm$ 0.015 & 0.927708          $\pm$ 0.066 & 0.771216            $\pm$ 0.148 \\ \hline
\multirow{3}{*}{\shortstack[l]{P.cysts \\ (w/o imp)}}  & prop. acc & 0.800569           $\pm$ 0.035 & 0.805143  $\pm$ 0.049 &                 0.793274 $\pm$ 0.069 & 0.824060          $\pm$ 0.019 & \textbf{0.828141            $\pm$ 0.012} \\  
                                    & acc       & 0.819493           $\pm$ 0.041 & 0.826413  $\pm$ 0.039 &                 0.814590 $\pm$ 0.060 & \textbf{0.842561          $\pm$ 0.004} & 0.824394            $\pm$ 0.033 \\  
                                    & kappa     & 0.756949           $\pm$ 0.054 & 0.764035  $\pm$ 0.052 &                 0.747127 $\pm$ 0.086 & \textbf{0.785441          $\pm$ 0.006} & 0.762919            $\pm$ 0.041 \\ \hline
\multirow{3}{*}{H.larvae}           & prop. acc & 0.954182           $\pm$ 0.008 & 0.964213  $\pm$ 0.017 &                 \textbf{0.964349 $\pm$ 0.012} & 0.941846          $\pm$ 0.039 & 0.951471            $\pm$ 0.014 \\  
                                    & acc       & 0.955450           $\pm$ 0.002 & 0.959558  $\pm$ 0.015 &                 \textbf{0.965561 $\pm$ 0.004} & 0.958926          $\pm$ 0.014 & 0.943128            $\pm$ 0.010 \\  
                                    & kappa     & 0.789743           $\pm$ 0.010 & 0.800052  $\pm$ 0.099 &                 \textbf{0.837948 $\pm$ 0.029} & 0.804689          $\pm$ 0.082 & 0.705475            $\pm$ 0.069 \\ \hline
\multirow{3}{*}{H.eggs}             & prop. acc & 0.936743           $\pm$ 0.011 & 0.936091  $\pm$ 0.005 &                 \textbf{0.937209 $\pm$ 0.008} & 0.931806          $\pm$ 0.007 & 0.930967            $\pm$ 0.006 \\  
                                    & acc       & \textbf{0.942634           $\pm$ 0.016} & 0.943938  $\pm$ 0.003 &                 0.942634 $\pm$ 0.009 & 0.908518          $\pm$ 0.022 & 0.853107            $\pm$ 0.025 \\  
                                    & kappa     & 0.899307           $\pm$ 0.027 & \textbf{0.901604  $\pm$ 0.006} &                 0.898922 $\pm$ 0.015 & 0.831488          $\pm$ 0.043 & 0.719695            $\pm$ 0.054 \\ \hline
\multirow{3}{*}{P.cysts}            & prop. acc & 0.732716           $\pm$ 0.056 & 0.769748  $\pm$ 0.026 &                 \textbf{0.780300 $\pm$ 0.018} & 0.748843          $\pm$ 0.048 & 0.744811            $\pm$ 0.068 \\  
                                    & acc       & 0.740973           $\pm$ 0.056 & 0.792755  $\pm$ 0.027 &                 \textbf{0.816905 $\pm$ 0.027} & 0.818066          $\pm$ 0.022 & 0.731104            $\pm$ 0.082 \\  
                                    & kappa     & 0.580626           $\pm$ 0.092 & 0.652254  $\pm$ 0.051 &                 \textbf{0.699603 $\pm$ 0.054} & 0.689325          $\pm$ 0.039 & 0.450283            $\pm$ 0.243 \\ \hline
\end{tabular}
\end{table*}

\section{Discussion}

\subsection{Does the confidence-based sampling improve the DeepFA looping?}
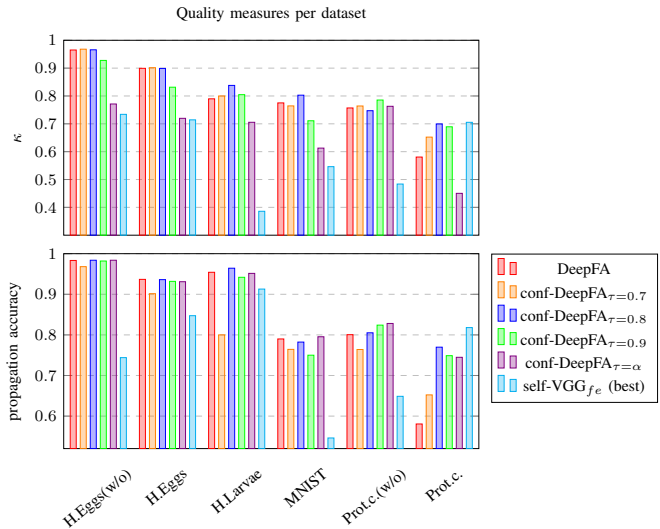
\begin{figure}[b]
    \begin{adjustbox}{width=0.5\textwidth}
    \begin{tabular}{c}
    \begin{tikzpicture}
        \begin{axis}[
            title={\footnotesize Quality measures per dataset},
            width  = \linewidth,
            height = 5cm,
            major x tick style = transparent,
            ticklabel style = {font=\footnotesize},
            label style = {font=\footnotesize},
            ybar,
            bar width=3pt,
            ymajorgrids = true,
            ylabel = {\footnotesize $\kappa$},
            symbolic x coords={H.Eggs(w/o), H.Eggs,H.Larvae, MNIST, Prot.c.(w/o),Prot.c.},
            xtick = data,
            xticklabel style={rotate=40},
            xticklabel=\empty,
            scaled y ticks = false,
            ymin=0.3,
            ymax=1,
            grid style=dashed,
            ytick = {0.4, 0.5, 0.6, 0.7, 0.8, 0.9, 1},
        ]
                
            \addplot[style={red,fill=red,mark=none,fill opacity=0.3}]
                coordinates {(H.Eggs(w/o),0.965085)  (H.Eggs,0.899307)  (H.Larvae,0.789743) (MNIST,0.775103)  (Prot.c.(w/o),0.756949) (Prot.c.,0.580626)};
            \addplot[style={orange,fill=orange,mark=none,fill opacity=0.3}]
                coordinates {(H.Eggs(w/o),0.968042)  (H.Eggs,0.901504)  (H.Larvae,0.800052) (MNIST,0.764348)  (Prot.c.(w/o),0.764035)  (Prot.c.,0.652254)};
            \addplot[style={blue,fill=blue,mark=none,fill opacity=0.3}]
                coordinates {(H.Eggs(w/o),0.965848)  (H.Eggs,0.898922)  (H.Larvae,0.837948) (MNIST,0.802863)  (Prot.c.(w/o),0.747127)  (Prot.c.,0.699603)};
            
            \addplot[style={green,fill=green,mark=none,fill opacity=0.3}]
                coordinates {(H.Eggs(w/o),0.927708)  (H.Eggs,0.831488)  (H.Larvae,0.804689) (MNIST,0.710961)  (Prot.c.(w/o),0.785441)  (Prot.c.,0.689325)};
                
            \addplot[style={violet,fill=violet,mark=none,fill opacity=0.3}]
                coordinates {(H.Eggs(w/o),0.771216)  (H.Eggs,0.719695)  (H.Larvae,0.705475) (MNIST,0.612766)  (Prot.c.(w/o),0.762919)  (Prot.c.,0.450283)};
            \addplot[style={cyan,fill=cyan,mark=none,fill opacity=0.3}]
                coordinates {(H.Eggs(w/o),0.734030)  (H.Eggs,0.714227) (H.Larvae,0.385892)  (MNIST,0.546162)  (Prot.c.(w/o),0.483706) (Prot.c.,0.705397)};
        \end{axis}
    \end{tikzpicture}\vspace{-5mm}\hspace{30mm}
    \\
    
    \begin{tikzpicture}
        \begin{axis}[
            width  = \linewidth,
            height = 5cm,
            major x tick style = transparent,
            ticklabel style = {font=\footnotesize},
            label style = {font=\footnotesize},
            ybar,
            bar width=3pt,
            ymajorgrids = true,
            ylabel = {\footnotesize propagation accuracy},
            symbolic x coords={H.Eggs(w/o), H.Eggs,H.Larvae, MNIST, Prot.c.(w/o),Prot.c.},
            xtick = data,
            xticklabel style={rotate=40},
            scaled y ticks = false,
            ymin=0.52,
            ymax=1,
            grid style=dashed,
            ytick = {0.6, 0.7, 0.8, 0.9, 1},
            legend pos=outer north east
        ]
                
            \addplot[style={red,fill=red,mark=none,fill opacity=0.3}]
                coordinates {(H.Eggs(w/o),0.983293)  (H.Eggs,0.936743)  (H.Larvae,0.954182) (MNIST,0.790000)  (Prot.c.(w/o),0.800569) (Prot.c.,0.580626)};
                \addlegendentry{\footnotesize DeepFA}
            \addplot[style={orange,fill=orange,mark=none,fill opacity=0.3}]
                coordinates {(H.Eggs(w/o),0.968042)  (H.Eggs,0.901504)  (H.Larvae,0.800052) (MNIST,0.764348)  (Prot.c.(w/o),0.764035)  (Prot.c.,0.652254)};
                \addlegendentry{\footnotesize conf-DeepFA$_{\tau=0.7}$}
            \addplot[style={blue,fill=blue,mark=none,fill opacity=0.3}]
                coordinates {(H.Eggs(w/o),0.983832)  (H.Eggs,0.936091)  (H.Larvae,0.964213) (MNIST,0.782286)  (Prot.c.(w/o),0.805143)  (Prot.c.,0.769748)};
                \addlegendentry{\footnotesize conf-DeepFA$_{\tau=0.8}$}
            
            \addplot[style={green,fill=green,mark=none,fill opacity=0.3}]
                coordinates {(H.Eggs(w/o),0.981945)  (H.Eggs,0.931806)  (H.Larvae,0.941846) (MNIST,0.750000)  (Prot.c.(w/o),0.824060)  (Prot.c.,0.748843)};
                \addlegendentry{\footnotesize conf-DeepFA$_{\tau=0.9}$}
                
            \addplot[style={violet,fill=violet,mark=none,fill opacity=0.3}]
                coordinates {(H.Eggs(w/o),0.983832)  (H.Eggs,0.930967)  (H.Larvae,0.951471) (MNIST,0.795429)  (Prot.c.(w/o),0.828141)  (Prot.c.,0.744811)};
                \addlegendentry{\footnotesize conf-DeepFA$_{\tau=\alpha}$}
            \addplot[style={cyan,fill=cyan,mark=none,fill opacity=0.3}] 
                coordinates {(H.Eggs(w/o),0.744004)  (H.Eggs,0.847308) (H.Larvae,0.912837)  (MNIST,0.546162)  (Prot.c.(w/o),0.648739) (Prot.c.,0.817978)};
                \addlegendentry{\footnotesize self-VGG$_{fe}$ (best)}
        \end{axis}
    \end{tikzpicture}

    \end{tabular}
    \end{adjustbox}

    \caption{Results of $\kappa$ (top) and propagation accuracies (bottom) for the studied datasets, considering self-VGG-16$_{fe}$ (best result) and DeepFA experiments. Our confidence-based DeepFA variations proposed in this paper are marked as \emph{DeepFA$_{\tau}$}. The datasets are ordered by higher $\kappa$ values in $x$ axis (from left to right).
    }
    \label{f.results_datasets}
    \end{figure}

Figure~\ref{f.results_datasets} shows the average $\kappa$ and propagation accuracy for the \emph{DeepFA} looping with fully-pseudo-labeling of all samples (\emph{DeepFA}), our proposed \emph{conf-DeepFA} using OPFSemi's confidence sampling for pseudo labeling, with different ways of selecting the confidence threshold $\tau$, and the best result for the VGG-16 experiments (\emph{self-VGG$_{fe}$}, see Sec.~\ref{ss.exps}), for all six studied datasets. For datasets yielding higher $\kappa$ values, we note that \emph{DeepFA} obtained similar results compared with our proposed \emph{conf-DeepFA} modifications. However, we see a gain of almost 5\% in $\kappa$ and propagation accuracy for the most challenging datasets. For \emph{P.cysts} with impurities, the gain is actually higher than $10\%$ in $\kappa$ and $17\%$ in propagation accuracy -- for which \emph{DeepFA} obtained worse results than VGG-16. In short, our proposal of using \emph{DeepFA} with OPFSemi's confidence sampling (\emph{conf-DeepFA}) entries in Fig.~\ref{f.results_datasets}) obtained the best results for most tested datasets.

\subsection{Does the confidence-based sampling improve DeepFA along the iterations?}
\begin{figure}[tb]
    \begin{adjustbox}{width=0.5\textwidth}
    \begin{tabular}{c}
    \begin{tikzpicture}
        \begin{axis}[
            title={\footnotesize Quality measures for 5 iterations on MNIST dataset},
            width  = 0.8\linewidth,
            height = 5cm,
            major x tick style = transparent,
            ticklabel style = {font=\footnotesize},
            label style = {font=\footnotesize},
            ybar,
            bar width=4pt,
            ymajorgrids = true,
            ylabel = {\footnotesize $\kappa$},
            symbolic x coords={1, 2, 3, 4, 5},
            xtick = data,
            xticklabel=\empty,
            scaled y ticks = false,
            ymin=0.53,
            ymax=1,
            grid style=dashed,
            ytick = {0.55,0.6, 0.65, 0.7, 0.75, 0.8, 0.85, 0.9, 0.95, 1},
        ]
                
            \addplot[style={red,mark=star,smooth}]
                coordinates {(1,0.649159)  (2,0.676172)  (3,0.69681) (4,0.735331)  (5,0.713007) };
                
            \addplot[style={orange,mark=otimes,smooth}]
                coordinates {(1,0.576352)  (2,0.671512)  (3,0.653695) (4,0.67825)  (5,0.725859) };
                
            \addplot[style={blue,mark=square,smooth}]
                coordinates {(1,0.665802)  (2,0.726419)  (3,0.793926) (4,0.764382)  (5,0.774711) };
            
            \addplot[style={green,mark=o,smooth}]
                coordinates {(1,0.571570)  (2,0.675535)  (3,0.733785) (4,0.693337)  (5,0.674328) };
                
            \addplot[style={violet,mark=triangle,smooth}]
                coordinates {(1,0.699107)  (2,0.767959)  (3,0.780475) (4,0.746952)  (5,0.680281) };
                
            \addplot[style={cyan, mark=diamond,smooth}]
                coordinates {(1,0.549358)  (2,0.549315)  (3,0.545571) (4,0.556706)  (5,0.558940) };
        \end{axis}
    \end{tikzpicture} \vspace{-5mm}\hspace{25mm}
    \\
    
    \begin{tikzpicture}
        \begin{axis}[
            width  = 0.8\linewidth,
            height = 5cm,
            major x tick style = transparent,
            ticklabel style = {font=\footnotesize},
            label style = {font=\footnotesize},
            ybar,
            bar width=4pt,
            ymajorgrids = true,
            ylabel = {\footnotesize propagation accuracy},
            symbolic x coords={1, 2, 3, 4, 5},
            xtick = data,
            scaled y ticks = false,
            ymin=0.53,
            ymax=1,
            grid style=dashed,
            ytick = {0.55,0.6, 0.65, 0.7, 0.75, 0.8, 0.85, 0.9, 0.95, 1},
            legend pos=outer north east
        ]
                
            \addplot[style={red,mark=star,smooth}]
                coordinates {(1,0.667714)  (2,0.711429)  (3,0.726) (4,0.755429)  (5,0.736571) };
                \addlegendentry{\footnotesize DeepFA}
                
            \addplot[style={orange,mark=otimes,smooth}]
                coordinates {(1,0.593714)  (2,0.669714)  (3,0.68) (4,0.712286)  (5,0.748857) };
                \addlegendentry{\footnotesize conf-DeepFA$_{\tau=0.7}$}
                
            \addplot[style={blue,mark=square,smooth}]
                coordinates {(1,0.633429)  (2,0.723143)  (3,0.786857) (4,0.817714)  (5,0.801143) };
                \addlegendentry{\footnotesize conf-DeepFA$_{\tau=0.8}$}
            
            \addplot[style={green,mark=o,smooth}]
                coordinates {(1,0.604857)  (2,0.664571)  (3,0.738286) (4,0.760857)  (5,0.745429) };
                \addlegendentry{\footnotesize conf-DeepFA$_{\tau:0.9}$}
                
            \addplot[style={violet,mark=triangle,smooth}]
                coordinates {(1,0.674857)  (2,0.765143)  (3,0.826286) (4,0.821714)  (5,0.802857) };
                \addlegendentry{\footnotesize conf-DeepFA$_{\tau=\alpha}$}
                
            \addplot[style={cyan, mark=diamond,smooth}]
                coordinates {(1,0.579429)  (2,0.579143)  (3,0.579143) (4,0.581714)  (5,0.581143) };
                \addlegendentry{\footnotesize self-VGG$_{fe}$ (best)}
        \end{axis}
    \end{tikzpicture}

    \end{tabular}
    \end{adjustbox}

    \caption{Results of $\kappa$ (top) and propagation accuracies (bottom) for the MNIST dataset in one split over $5$ iterations, considering self-VGG-16$_{fe}$ (best result), \emph{DeepFA}, and \emph{conf-DeepFA} experiments.
    }
        \label{f.results_mnist}
    \end{figure}
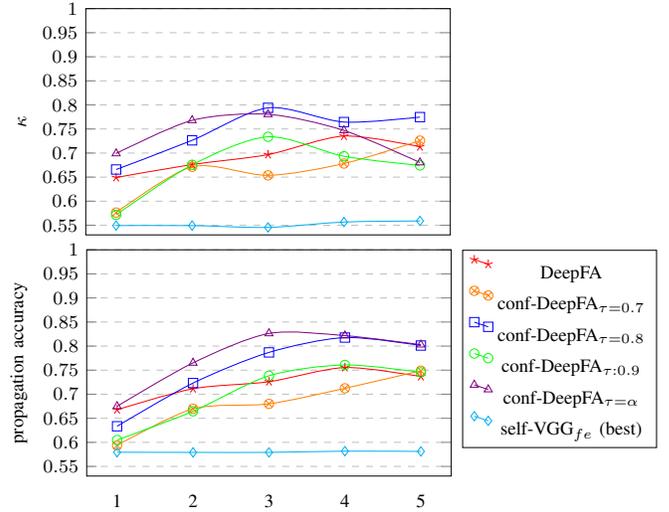

\begin{figure}[!ht]
    \centering
    \includegraphics[width=\linewidth]{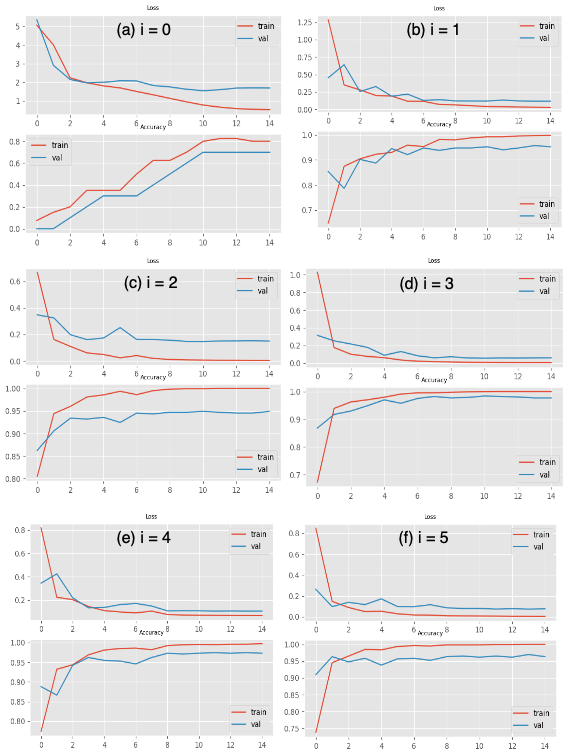} 
    \caption{Plots for loss and accuracy for one split of MNIST dataset. The (a) initial learning curves and the for each iteration (b, c, d, e, f) is presented.}
    \label{f.learningcurves}
\end{figure}

Figure~\ref{f.results_mnist} shows $\kappa$ and propagation accuracy for one split of MNIST  along five iterations of the proposed experiments. First, we see that all compared approaches yielded an increase from the first to the second iteration, except \emph{self-VGG-16$_{fe}$}. Also, we see that both $\kappa$ and the propagation accuracy slightly decrease after the third iteration. This may suggest that the proposed method saturates, mainly by the higher decrease in $\kappa$ despite of propagation accuracy. The learned pseudo-labels and the original images can be used as input for a better (known) deep architecture. Figure~\ref{f.learningcurves} shows the plot for train and validation loss and accuracy considering 20\% (from the $S$ set) as validation set during one split of MNIST training. The initial learning curve and the learning curves for each iteration are also shown. The learning curves show that the labeled samples can improve the network convergence along the iterations. A different deep network can be tested at the final stage. Also, some unsupervised quality measure can be proposed to define the best feature space found at certain iterations and, consequently, the best iteration of the method.

\subsection{Choosing OPFSemi's confidence threshold}
The proposed adaptive selection of the confidence threshold (\emph{conf-DeepFA$_{\tau=\alpha}$}) seems to be promising only for one of the tested datasets. It shows a higher decreasing in $\kappa$, when compared with the experiments without changing the confidence threshold $\tau$ along the iterations. As outlined in Sec.~\ref{ss.exps},  choosing OPFSemi's confidence value may depend on the dataset, its difficulty, number of samples, number of classes, and class imbalance. This can also be seen in Fig.~\ref{f.results_datasets} where it is not possible to define a \emph{single} confidence threshold $\tau$ for \emph{all} chosen datasets. Although this fact has been already noted in~\cite{Benato:2021}, it was not considered within a looping of data annotation as we proposed in this work. Rather, in~\cite{Benato:2021}, the authors proposed user interaction to define the best confidence value based on the user analysis of the 2D space projection guided by the data distribution and OPFSemi's confidence values (mapped to colors). We intend to follow the same strategy to find the best confidence value for \emph{conf-DeepFA} looping.

\subsection{Limitations}

As discussed before, we intend to explore other deep learning architectures to understand why and when the proposed \emph{conf-DeepFA} looping stagnates during network training. Also, we validated our method for only six datasets, one semi-supervised classifier, and one projection method. More experiments involving additional datasets, classifiers, and 2D projection techniques are needed to generalize our findings.

\section{Conclusion}
We proposed an approach for increasing the quality of image classification and extracted feature spaces when using very few supervised samples during the training process. We evaluate the feature space generated by VGG-16 by feature extraction and fine-tuning strategies when using the small number of supervised images available. We use the best resulting feature space to let the OPFSemi propagation technique label unsupervised samples on a 2D t-SNE projection of the feature space in an iterative fashion. To improve label propagation accuracies and  classification results, we include a confidence sampling strategy to OPFSemi's pseudo labeling to define the most confident samples for training VGG-16.

Our results show that the VGG-16 without fine-tuning, i.e., only being used for feature extraction, can improve accuracy and $\kappa$ with few supervised samples. Also, when considering OPFSemi with the confidence sampling strategy, our results show an improvement in propagation accuracy and $\kappa$ for most of the evaluated datasets. The small gain for some chosen confidence thresholds $\tau$ and some datasets shows that this choice may depend on the dataset. To solve this dataset dependency, we plan next to include user knowledge to provide a semi-automatic pseudo-labeling along the lines in\,\cite{Benato:2021} but considering the proposed looping of the deep feature annotation method. Additionally, we aim to explore more  datasets, as well as compare our proposed method with recent semi-supervised strategies for creating pseudolabels.


\section*{Acknowledgments}
The authors acknowledge FAPESP grants $\#2014/12236-1$, $\#2019/10705-8$, CAPES grants with Finance Code 001, and CNPq grants $\#303808/2018-7$.



\bibliographystyle{IEEEtran}
\bibliography{ref}
%
%


\end{document}